\documentclass[sigconf]{acmart}

\renewcommand\footnotetextcopyrightpermission[1]{} 
\usepackage{lipsum}
\usepackage{amsfonts}
\usepackage{mathtools}
\usepackage[linesnumbered, ruled]{algorithm2e}

\providecommand{\DontPrintSemicolon}{\dontprintsemicolon}
\usepackage{graphicx}
\usepackage{float}
\usepackage{tabularx}
\usepackage{multirow}
\usepackage{threeparttable}
\usepackage{csvsimple}
\usepackage[caption=false,font=footnotesize]{subfig}
\usepackage{amsmath}
\usepackage{tikz}
\usetikzlibrary {arrows.meta,automata,positioning,shadows}
\usepackage{pgfplots}
\pgfplotsset{compat=1.16}
\AtBeginDocument{%
  \providecommand\BibTeX{{%
    \normalfont B\kern-0.5em{\scshape i\kern-0.25em b}\kern-0.8em\TeX}}}

\begin{document}

\title{Learning Traffic Signal Control via Genetic Programming}

\author{Xiao-Cheng Liao}
\orcid{0009-0001-6992-8339}
\author{Yi Mei}
\orcid{0000-0003-0682-1363}
\author{Mengjie Zhang}
\orcid{0000-0003-4463-9538}
\affiliation{%
  \institution{Victoria University of Wellington}
  \department{Centre for Data Science and Artificial Intelligence \& School of Engineering and Computer Science}
  \city{Wellington}
  \country{New Zealand}
}
\email{{xiaocheng, yi.mei, mengjie.zhang}@ecs.vuw.ac.nz}

\renewcommand{\shortauthors}{Liao, et al.}

\begin{abstract}
The control of traffic signals is crucial for improving transportation efficiency. 
Recently, learning-based methods, especially Deep Reinforcement Learning (DRL), garnered substantial success in the quest for more efficient traffic signal control strategies.
However, the design of rewards in DRL highly demands domain knowledge to converge to an effective policy, 
and the final policy also presents difficulties in terms of explainability.
In this work, a new learning-based method for signal control in complex intersections is proposed.
In our approach, we design a concept of phase urgency for each signal phase.
During signal transitions, the traffic light control strategy selects the next phase to be activated based on the phase urgency.
We then proposed to represent the urgency function as an explainable tree structure. 
The urgency function can calculate the phase urgency for a specific phase based on the current road conditions.
Genetic programming is adopted to perform gradient-free optimization of the urgency function.
We test our algorithm on multiple public traffic signal control datasets. 
The experimental results indicate that the tree-shaped urgency function evolved by genetic programming outperforms the baselines, including a state-of-the-art method in the transportation field and a well-known DRL-based method.
Our code is available online\footnote{https://github.com/Rabbytr/gplight}.

\end{abstract}




\maketitle

\section{Introduction}
\label{sec:introduction}

Traffic signals coordinate traffic flow with different directions at signalized intersections, and play an important role in enhancing both transportation efficiency and road safety \cite{liao2022combining}.
Ineffective traffic signal plans lead to wasted time for commuters on the roads. 
Most current traffic signal control systems are not based on decisions made according to the dynamic traffic environment.
For example, Sydney Coordinated Adaptive Traffic System \cite{lowrie1990scats}, designed based on pre-determined cycle time plan, is still widely adopted at real signalized intersections around the world.

With advancements in deep learning \cite{lecun2015deep} and the accessibility of transportation infrastructure (e.g., surveillance cameras, road sensors and internet of vehicles) \cite{ji2020survey}, there is an emerging trend \cite{chen2020toward,wei2019colight, wei2019presslight} of utilizing Deep Reinforcement Learning (DRL) to address Traffic Signal Control (TSC) problem.
DRL methods can seek effective traffic signal control strategies based on feedback from the environment. 
These strategies can dynamically adjust traffic signals according to the real-time conditions at intersections and have demonstrated superior performance when compared to traditional transportation methods \cite{haydari2020deep}.

Despite promising results achieved in TSC, the existing DRL methods still suffer from two major problems.
One is the intricate design of essential components, with the reward system being particularly demanding.
Designing a reasonable reward often requires a significant amount of domain knowledge and expertise \cite{wei2019presslight}. 
Otherwise, it is easy to encounter issues such as overly delayed rewards and credit assignment over long time scales \cite{salimans2017evolution}.
For instance, within a time step, it is challenging to describe the effectiveness of an action (i.e., transitioning from one traffic signal phase to another) using an immediate reward.

The other issue is that the ultimate signal control policy learned by DRL is typically based on a complex deep neural network. 
This complexity impedes human or expert understanding, making it difficult to explain the learned policy.
However, explainable traffic signal control strategies are crucial \cite{zhu2022extracting}.
The absence of transparency poses significant barriers to establishing trust from users and for dispatchers to examine potential weaknesses in the policies \cite{sheh2017did,mei2022explainable}.
In addition, drivers should also be able to anticipate the next traffic light change while waiting for a green signal; otherwise, it may lead to traffic confusion.
For example, drivers in the through lane may forcefully enter the right-turn lane because they cannot anticipate the arrival of a green signal.

To address the shortcomings mentioned above, this paper proposes a traffic signal control optimization algorithm based on genetic programming (GP) \cite{liao2023towards,liao2023uncertain}.
The proposed algorithm, called GPLight, aims to evolve an effective urgency function.
This urgency function is an explicit tree expression that can evaluate the priority for each traffic light phase at an intersection in real time.
Each time a transition in traffic signal phases is necessary at an intersection, the urgency function considers the traffic movement features on the lanes that can be affected by the phase, and generates an urgency value for that specific phase.
This outputted phase urgency can be considered as the priority for a green light demand at the current intersection state. Consequently, the phase with the maximum phase urgency will be selected as the next phase.
The proposed GPLight treats the traffic signal control as a black-box model and conducts a global search for the most accurate urgency function.
This not only eliminates the necessity for reward design grounded in domain knowledge, but also maintains explainability in the final signal control strategy.
The main contributions of this paper are as follows:
\begin{enumerate}
    \item To the best of our knowledge, we are among the first to adapt GP for traffic signal control in the complete 8-phase multi-intersection scenarios.

    \item The performance of the proposed method was compared with a well-known reinforcement learning method, MPLight \cite{chen2020toward}, on publicly available real-world datasets. The experimental results indicate that our approach significantly outperforms MPLight as well as state-of-the-art heuristic-based method \cite{varaiya2013max} in most scenarios.

    \item Both high-quality and human-understandable traffic signal control strategies are achieved by tree-like structured expressions.

\end{enumerate}

The rest of this paper is organized as follows.
In Section \ref{sec:background}, we review the prior studies concerning traffic signal control and introduce the traffic signal control problem.
We presents the proposed GPLight in Section \ref{sec:method} and validate its effectiveness and advantages in Section \ref{sec:experiments}.
Finally, we conclude in Section \ref{sec:conclusion}.

\section{Background}
\label{sec:background}

\subsection{Problem Model}
\label{sec:problem}
In this section, some important definitions related to the traffic signal control to be solved in this paper is introduced. 

\begin{figure}[!t]
\centering
\includegraphics[width=\columnwidth,trim=0 0 0 0,clip]{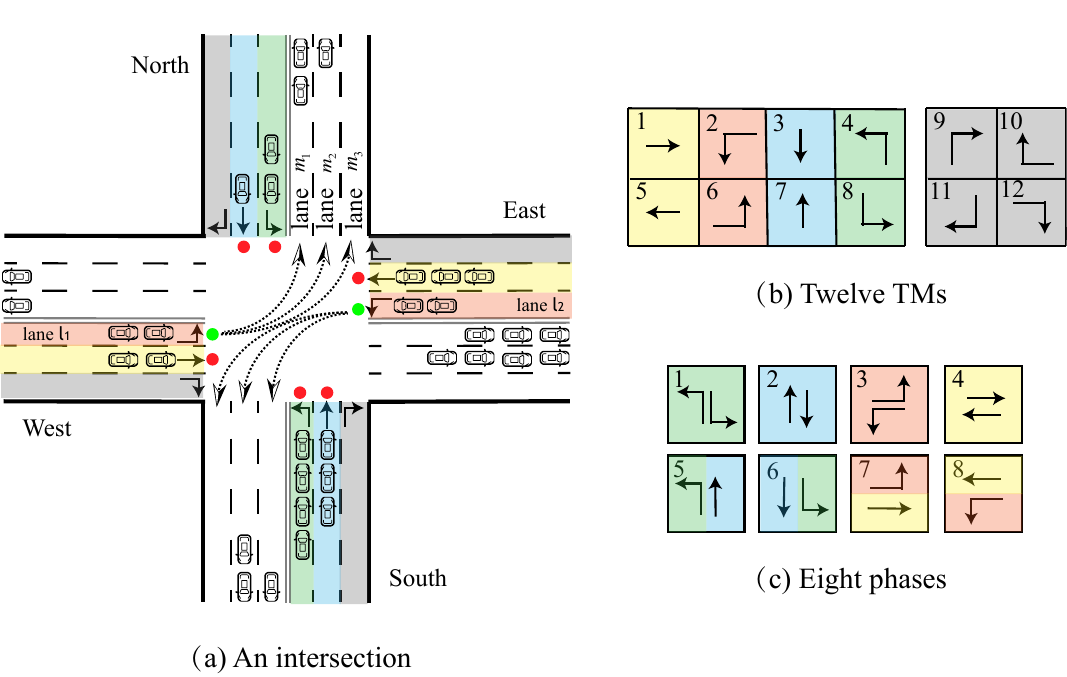}
\caption{Example of an intersection, traffic movements, and different phases. In this figure, phase $s_3$ is currently activated.}
\label{fig_example}
\end{figure}

\textit{Definition 1 (Intersection structure)}. An example of the intersection structure is shown in Figure \ref{fig_example} (a). It consists of eight roads which can be categorized into two types: incoming roads and outgoing roads. 
Each road is composed of three lanes, one turning left, one going straight, and one turning right.
In total, there are twelve incoming lanes and twelve outgoing lanes in an intersection. 
A vehicle arriving on incoming lanes $l$ can cross the intersection and move to one of the corresponding outgoing lanes $m$. 

\textit{Definition 2 (Traffic movement (TM))}. 
Each traffic movement represents vehicles crossing an intersection in a particular direction, including left, right or straight. 
We assume that traffic must travel on the right side and U-turns are not allowed. 
As shown in Figure \ref{fig_example} (b), there are twelve traffic movements, including four go-straight TMs, four left-turn TMs and four right-turn TMs (gray directions). Among them, right turn TMs are permitted at all times. Therefore, typically, there are eight adjustable TMs.

\textit{Definition 3 (Movement signal and signal phase)}. A movement signal refers to the movement of traffic in each direction in an intersection. Specifically, the green signal indicates that traffic in a particular direction is allowed to move safely, while red and yellow signals signify that traffic in that direction is prohibited from moving. 

A traffic signal phase encompasses a set of allowed traffic movements.
Some signals cannot turn `green' at the same time, and there are eight possible pairs of non-conflicting signals at an intersection, which are called signal phases, $S = \{s_1,s_2,...s_8\}$.
At one time, only one phase can be activated at an intersection. 
In Figure \ref{fig_example}, phase $s_3$ is activated, the traffic from $l_1$ and $l_2$ is allowed to turn left to corresponding possible outgoing lanes.

In this work, we assume that each road is always composed of three lanes: left-turn lane, go-straight lane, and right-turn lane.
From Figure \ref{fig_example}, it can be observed that each phase includes two incoming lanes $l_1$ and $l_2$, each with its own three downstream lanes $\{m_1, m_2, m_3\}$.
Thus, for each $s_i \in S$, $|s_i| = 2 \times 3 = 6$.

\textit{Definition 4 (Traffic Network)}. A traffic network comprises a collection of intersections $(I_1,I_2,...,I_N)$ connected by a set of roads $(R_1,R_2,...R_M)$, where $N$ represents the number of intersections, $M$ denotes the number of roads.

The summary of notations is outlined in Table \ref{table_notations}.

\textit{\textbf{Objective:} (Average travel time)}. A traffic network consists of intersections and roads. Given the time period of analysis, each intersection is controlled at each step $t$, making an optimal decision to choose appropriate signal phase by observing the current traffic conditions at the intersection. The objective is to minimize the average travel time for all vehicles spent between entering and leaving area in the network.

\begin{table}[!t]
\caption {Summary of notations}
\centering
\begin{tabular}{c>{\arraybackslash}p{6cm}}
\hline
\hline
Notations      & Meaning  \\ 
\hline
$l$   & incoming lanes for an intersection \\ 
$m$ & outgoing lanes for an intersection  \\ 
$s$ & a traffic signal phase in an intersection \\ 
$S$ & a set of possible signal phases in an intersection \\
$I_1,I_2,..,I_N$ & intersections in a traffic network \\ 
$R_1,R_2,...,R_m$ & a set of roads in a traffic network \\ 
$w(l)$ & number of waiting vehicles on lane $l$ \\
$x(l)$ & total number of vehicles on lane $l$ \\  
\hline
\hline
\end{tabular}
\label{table_notations}
\end{table}

\subsection{Related Work}
Numerous traffic signal control approaches have been proposed, categorizable into two typical groups: traditional methods, and learning-based methods.

\subsubsection{Traditional Traffic Signal Control Methods}
Traditional methods for traffic signal control can be further categorized into four types.

\textit{Fixed time control.}
Webster et al. \cite{webster1958traffic} introduced the fixed-time control approach in 1958, which entails modifying signal phases according to predefined rules specified in signal plans.
The fixed-time method controls the switching of traffic signal phases in a very mechanical sequence, without considering states of vehicles on different lanes.
This could easily lead to highly imbalanced traffic flow.
However, due to its simplicity, it is widely adopted in the real world.

\textit{Actuated methods.}
Some studies \cite{fellendorf1994vissim, mirchandani2001real} established a set of rules, and the activation of the traffic signal is contingent upon adherence to these predefined rules alongside real-time data. For instance, a rule could stipulate that the green signal is assigned to a specific traffic movement only when the queue length exceeds a designated threshold.

\textit{Selection-based adaptive control methods.}
In later developments, some methods and systems \cite{koonce2008traffic}, such as SCATS \cite{lowrie1990scats} and SCOOT \cite{hunt1982scoot}, 
coordinate traffic signals based on a set of signal plans that are designed manually.
In these approaches, a series of signal plans are predefined, and then, based on the real-time traffic conditions of roads, the system selects which signal plan to adopt.

\subsubsection{Optimization-based methods}
Traditional methods for traffic signal control heavily depend on human expertise, as they necessitate the manual design of traffic signal plans or rules.
It also lacks integration with an optimization process, leading to potential performance shortcomings.
Classical optimization-based methods typically involve optimizing travel time by assuming a uniform arrival rate \cite{webster1966traffic,roess2004traffic}. 
Subsequently, a traffic signal plan, encompassing cycle length and phase ratios, can be computed using formulas based on traffic data.
Some researchers utilize meta-heuristics \cite{liao2023crowd}, such as genetic algorithms \cite{li2018signal,tung2014novel}, differential evolution \cite{cakici2019differential, baskan2019multiobjective}, particle swarm optimization \cite{jia2019multi, chuo2017evolvable} and ant colony optimization \cite{baskan2011ant, renfrew2012traffic}, to optimize signal plans.
The signal type, cycle time, signal offset and green time are typically used as decision variables.
Optimization-based methods depend less on human knowledge, determining traffic signal plans based on observed traffic data, and demonstrated promising results.
However, these methods still rely on pre-optimized signal plans, making it challenging to handle dynamic and uncertain traffic conditions.

Varaiya proposed the Max-Pressure (MP) \cite{varaiya2013max} method that does not rely on predefined signal plans and can adjust traffic signals based on real-time traffic conditions at intersections.
MP is a state-of-the-art method in the transportation field \cite{wei2019presslight} and its primary objective is to maximize network throughput, thereby minimizing travel time.
However, it still relies on strong assumptions, such as assuming unlimited capacity of downstream lanes to simplify traffic conditions. 
These assumptions may limit the effectiveness of the MP method in real-world scenarios, as they may not always hold true in actual conditions.

\subsubsection{Learning for Traffic Signal Control}
Different from traditional methods, learning-based approaches do not require the predefined or pre-optimized static signal plans and do not make strong assumptions about traffic data.
It can learn directly from intersections with feedback from the transportation system without prior knowledge about a given environment.

At present, the most prominent learning-based method for traffic signal control is DRL.
In DRL, one approach is to centrally control signals in all intersections of the road network \cite{prashanth2010reinforcement}.
This involves the agent directly determining actions for all intersections, but mastering the task is challenging due to the curse of dimensionality in the action space.
Some studies, such as \cite{kuyer2008multiagent} and \cite{van2016coordinated}, have explored the use of a multi-agent DRL method to jointly model two adjacent intersections, employing centralized global joint actions.
However, these approaches encounter scalability issues during deployment. 
Consequently, as the network scale expands, centralized optimization becomes infeasible due to the combinatorially large joint action space, impeding the widespread adoption of this method for city-level control.

To mitigate this issue, the method of modeling each intersection as an individual agent was proposed \cite{wiering2000multi}.
In these methods, each intersection is controlled by a separate agent \cite{wei2019colight,wei2019presslight,wei2018intellilight}, and in some studies \cite{zheng2019learning,chen2020toward}, these agents share parameters.
The state representing the quantitative description of the traffic condition at that intersection.
The action corresponds to the traffic signal, and the reward serves as a measure of transportation efficiency, such as delay \cite{el2010agent}, queue length \cite{wei2019colight}, pressure \cite{wei2019presslight}.
However, these methods typically require designing a reward that may not be directly related to the problem objective, and they often demand a certain level of domain knowledge.

In addition to DRL-based methods, Ricalde and Banzhaf \cite{ricalde2016genetic, ricalde2017evolving} undertook some preliminary work using GP with an epigenetics design in the TSC domain.
This is the most similar work to the approach presented in this paper.
Nevertheless, their method operates in scenarios with only vertical and horizontal traffic flows, similar to a 2-phase setup, which is uncommon in real-life situations.
Their approaches cannot be applied to the complete 8-phase multi-intersection scenarios.
The utilization of GP to develop effective and explainable traffic signal control strategies still encounters limitations that necessitate further advancements.

\section{GPLight}
\label{sec:method}

\begin{figure*}[!t]
\centering
\includegraphics[width=\textwidth,trim=0 0 0 0,clip]{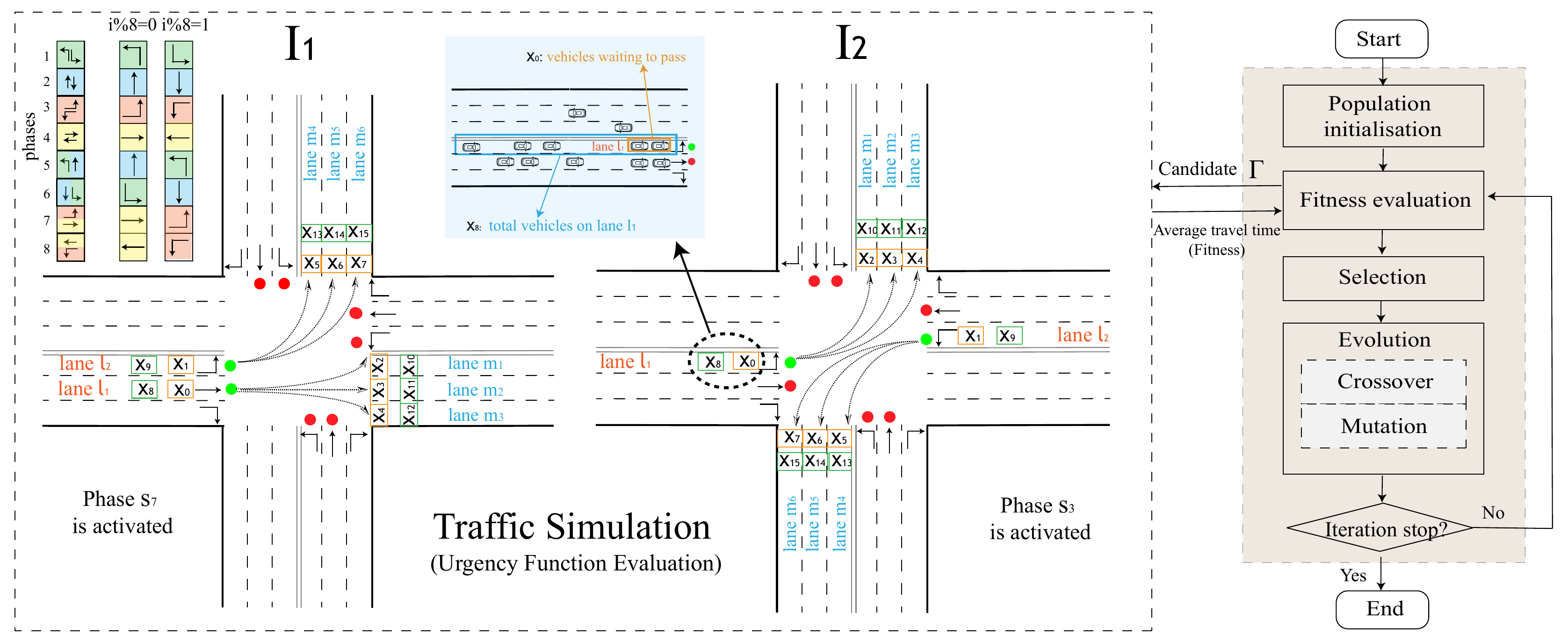}
\caption{Framework of the proposed GPLight. }
\label{fig_gpphase}
\end{figure*} 

In this section, we describe the proposed method, namely GPLight, which uses GP to learn traffic signal control.
Specifically, we first present the detailed design of the urgency function that is used to calculate the phase urgency for each phase at an intersection.
Then, we describe the process of determining next signal phase for each intersection at every decision point (i.e., each step $t$) based on the urgency function.
Afterwards, the evolutionary process of the proposed GPLight for evolving the urgency function is described, including the framework, operators of GP and simulation-based fitness evaluation.

\subsection{Urgency Function}
\label{sec:urgency_func}

In this work, we introduced the concept of \textit{phase urgency}, which can measure the urgency of activating a green light for a specific phase.
Our objective is to utilize the urgency function to calculate the urgency value for each phase. 
At each stage of traffic signal transitions at an intersection, the signal light phase with the highest urgency value, signifying it as the most urgent, should be activated.
Through this approach, we aim to choose the signal phase with the highest urgency at each signal transition, thereby reducing waiting times for vehicles at intersections and effectively coordinating traffic flows from different directions.

Accordingly, the crucial aspect lies in designing a suitable urgency function.
Roads associated with different phases are distinct, and the lanes associated with them are heterogeneous, potentially encompassing incoming, outgoing, left-turn, and go-straight lanes.
Therefore, the urgency function should be capable of accurately and effectively calculating the urgency for each phase, adapting to diverse situations occurred over time.
In the following, we progressively introduce the design of the urgency function.

\subsubsection{Lane Features} 
\label{sec:lane_features}
With the advancement of vehicular ad-hoc network \cite{al2014comprehensive} technology and continuous improvements in road infrastructure, enormous real-time information on key vehicle and road conditions can be extracted and analyzed.
This work employs two features associated with lane, which also were widely utilized in previous research \cite{chen2020toward,zhang2022expression}.
For each lane, two pieces of information from the lane are utilized by the urgency function, including the number of waiting vehicles $w(l)$ on lane $l$, and the total number of vehicles $x(l)$ on lane $l$, where $x(l)$ both includes vehicles in motion and waiting vehicles. 
Since one phase involves two non-conflicting traffic movements, for any given traffic movement, vehicles from the incoming lane can move to three outgoing lanes, including left-turn, go straight and right-turn outgoing lanes.
Therefore, in an intersection, each phase involves eight lanes, including two incoming lanes and six outgoing lanes. 
As shown in Figure \ref{fig_gpphase}, phase $s_7$ of intersection $I_1$ and phase $s_3$ of intersection $I_2$ are activated, with the incoming lanes ($l_1$, $l_2$) represented in purple font and the outgoing lanes ($m_1$,$m_2$,...$m_6$) in blue font.

\subsubsection{Phase Inputs for Urgency Function}
As aforementioned in section \ref{sec:lane_features}, two features $w(l)$ and $x(l)$ are available for each lane $l$.
Then, each road consists of three lanes, and each phase involves four roads. 
Therefore, the calculation of urgency for each phase will involve $2\times 3\times 4 = 24$ features.
To minimize the input dimensionality of the urgency function, in this work, we choose to exclude features from two right-turn lanes of incoming roads and their respective three downstream lanes, which are often included as inputs in the Q-network of reinforcement learning \cite{chen2020toward}.
Our rationale for this is that the TM of right-turn lanes is not influenced by the traffic signal.
Its impact on the calculation of urgency values might be relatively marginal.
Hence, the number of lanes that need to be taken into consideration in each phase is reduced to 8, and the total number of variables is decreased to 16.

For each phase, we can represent its features using 16 features gathered from 8 relevant lanes:
\begin{equation}
    X = \left\{x_0,x_1,....,x_{15}\right\}.
\end{equation}
A challenge of the feature encoding is that 
the lane linked to a $x_i \in X$ undergoes variations when urgency is computed for different phases.
For example, in one phase, $x_i$ ($i<8$) is defined to represent the number of waiting vehicles $w(l_j)$ in a specific lane $l_j$ during that phase. However, in different phases of the intersection, $x_i$ may represent the number of waiting vehicles $w(l_k)$ in another lane $l_k$, where $j \neq k$. 
This variation is the fundamental principle to base that the urgency function can calculate unique urgency values for different phases. 
However, it also brings some challenges to the design of $X$.
For an extreme example of poorly designed $X$, $x_i$ represents the number of waiting vehicles $w(l)$ on an incoming lane when calculating phase urgency of $s_1$, it might represent the total number of vehicles $x(l)$ on an outgoing lane when calculating phase urgency of $s_2$.
The contextual dependence in indicates that $x_i$ has a variable semantics across different phases, and its specific meaning is contingent upon the changes in phases.
To address this issue, we need to thoroughly consider such variations in the intersection control to ensure accurate modeling and understanding of intersection behavior.

To better distinguish each feature, in this paper, we adopt the following principle: the first eight features represent the number of waiting vehicles on lanes, while the subsequent eight features represent the total number of vehicles on lanes. 
Specifically, feature sets $\{x_i \mid i\equiv 0 \pmod{8}\}$ and $\{x_i \mid i\equiv 1 \pmod{8}\}$ come from the two incoming lanes $l_1$ and $l_2$, respectively.
We specify the ordering of the two incoming lanes for each phase, as shown in the top-left corner of Figure \ref{fig_gpphase}.
Feature sets $\{x_i \mid i\equiv 2 \pmod{8}\}$, $\{x_i \mid i\equiv 3 \pmod{8}\}$, and $\{x_i \mid i\equiv 4 \pmod{8}\}$ correspond to downstream left-turn, through, and right-turn lanes of incoming lane $l_1$, respectively. 
Feature sets $\{x_i \mid i\equiv 5 \pmod{8}\}$, $\{x_i \mid i\equiv 6 \pmod{8}\}$, and $\{x_i \mid i\equiv 7 \pmod{8}\}$ are derived from downstream left-turn, through, and right-turn lanes of incoming lane $l_2$, respectively.

Taking the example of the activated phase $s_7$ at intersection $I_1$ and the activated phase $s_3$ at intersection $I_2$, as illustrated in Figure \ref{fig_gpphase}, the specific representation of each feature is marked in the figure.
In phase $s_7$, the number of waiting vehicle in the left-turn outgoing lane to the east direction (represented as lane $m_1$) is defined as $x_2$, while in phase $s_3$, the waiting vehicles in the left-turn outgoing lane to the north direction represents feature $x_2$.

To simplify representation, we encapsulate the above feature extraction into a function $\mathcal{G}$:
\begin{equation}
    X_i = \mathcal{G}\left( s_i \right),
    \label{eq_preprocess}
\end{equation}
where $X_i$ represents the ordered input features for phase $s_i$. 
The urgency function can take $X_i$ as input and directly output the urgency value for phase $s_i$.

\subsubsection{Signal Control via Urgency Function}

Considering that each phase involves different incoming and outgoing lanes, and the traffic conditions vary in each lane, with many vehicles waiting on some lanes while others have only a few vehicles.
Therefore, by assessing the traffic flow conditions on lanes associated with each phase, calculating the urgency value for each phase, and selecting the phase with the highest priority, is an intuitive approach.
When a phase transition is required at any intersection, the urgency function can be utilized to calculate the next phase that is most appropriate for activation, as specified by Eq. \eqref{eq_selection}.
\begin{equation}
    s^* = \mathop{\arg\max}\limits_{s_i \in S}\ \left(\Gamma\left( \mathcal{G}\left(s_i\right) \right)\right)
    \label{eq_selection}
\end{equation}
where $\Gamma(\cdot)$ represents the urgency function, which is a tree-based function partially composed of features for each phase.
With the given urgency function, when a phase switch is required at an intersection, the currently most urgent phase can be determined.

\definecolor{myorange}{rgb}{0.91910675, 0.58281075, 0.43899817}
\definecolor{mygreen}{rgb}{0.48942421, 0.72854938, 0.56751036}
\begin{figure}[!t]
\centering
\begin{tikzpicture}
[level distance=10mm,
ternode/.style={fill=myorange!80, circle,inner sep=0pt,minimum size=5mm},
funnode/.style={fill=mygreen, circle,inner sep=0pt,minimum size=5mm},
level 1/.style={sibling distance=32mm},
level 2/.style={sibling distance=18mm},
level 3/.style={sibling distance=10mm},
my lable/.style={
rectangle split, rectangle split parts=#1, draw,rectangle split horizontal,anchor=center}
]
\node [funnode] {$+$} 
child { 
    node [funnode] {$-$}
    child {
        node [ternode] {$x_0$}
    }
    child { 
        node [funnode] {$\div$}
        child {
            node [ternode] {$x_1$}
        }
        child {
            node [ternode] {$x_2$}
        }
    }
}
child { 
    node [funnode] {$\times$}
    child {
        node [ternode] {$x_3$}
    }
    child { 
        node [ternode] {$x_4$} 
    }
};
\end{tikzpicture}
\caption{Tree representation of an urgency function $\Gamma(\cdot) = x_0 - x_1/x_2 + x_3x_4$.}
\label{fig_gptree}
\end{figure}
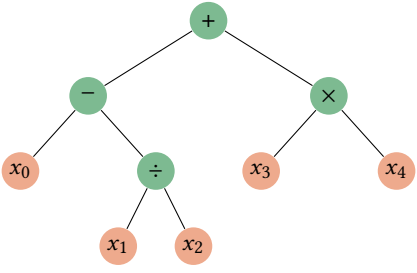 

\subsection{Evolutionary Process}

In this paper, we use genetic programming to evolve the urgency functions, which we called GPLight.
In GPLight, each individual is a tree-based function \cite{chen2023heuristic} and represents a candidate urgency function $\Gamma(\cdot)$.
An example of individual is illustrated in Figure \ref{fig_gptree}.
This tree-like individual is composed of terminal nodes on the leaf nodes and function nodes.
The terminal set is designed as the aforementioned features of a single phase:
\begin{align}
\begin{aligned}
&X_i = \{ x_0, x_1, x_2, x_3, x_4, x_5, x_6, x_7, \\
x_{8+0}, &x_{8+1}, x_{8+2}, x_{8+3}, x_{8+4}, x_{8+5}, x_{8+6}, x_{8+7} \},
\end{aligned}
\end{align}
where $X_i$ is the terminal set used in this work, wherein $x_{8+0}$ is equivalent to $x_{8}$, $x_{8+1}$ is equivalent to $x_{9}$ and so forth. 
Through this representation, it can be easily discerned that the features $x_{8i+k}$ and $x_{8j+k}$ originate from the same lane.
The function set is set as \eqref{eq_func_set}.
\begin{equation}
\left\{+,-,\times,\div,\min,\max\right\}.
\label{eq_func_set}
\end{equation}
Each function operates on two arguments, with the "$\div$" function ensuring protected division and returning one in case of division by zero. 
The $max$ and $min$ functions take two arguments, returning the maximum and minimum values, respectively.

The overall framework of GPLight for traffic signal control is presented in Figure \ref{fig_gpphase}.
The algorithm begins with the random generation of individuals by applying the ramped-half-and-half method \cite{luke2001survey}, thereby forming the initial population.
Then the evolutionary process starts.
Firstly, each individuals are compiled to urgency functions and each urgency function is evaluated based on simulation, which is presented in Algorithm \ref{alg_simulation}.

The simulation begins with the loading of a traffic dataset and initialization of the time step (lines 1-2). 
The algorithm then enters a main loop, where, for each time step $t$, it iterates over each intersection in the road network. 
At each intersection, real-time data is collected from lanes, preprocessed to obtain relevant information for each signal phase, and the most urgent phase is determined using the urgency function (lines 5-7).
The signal phase of each intersection is set to the most urgent phase, and the simulation progresses by switching red and yellow lights based on signal phase changes. 
Vehicle movement is simulated until the next signal phase transition, which occurs at the subsequent time step.
In this work, each time step includes a 10-second period for vehicle passage at each intersection, with a 3-second yellow light time and a 2-second red light time.
This iterative process continues for multiple time steps until the current time reaches the predetermined simulation duration. 
After the simulation completes, the average travel time of vehicles can be calculated, providing an important metric for evaluating the overall performance of an urgency function.

After the evaluation of all individuals is completed, promising individuals are selected by the tournament selection operator based on their fitness values.
Then, all individuals in the population undergo two genetic operators, including crossover and mutation.
In this work, we employ the subtree crossover operator, where the subtrees of two parents are randomly selected and exchanged to generate new individuals for the next generation. 
In addition, in the mutation operator, a randomly sampled subtree of the mutated individual is replaced by a newly generated subtree. 
The process is repeated until the maximum number of generations is reached, and the minimum vehicle travel time and the best urgency function $\Gamma^*(\cdot)$ can be returned.

\begin{algorithm}[!t]
\DontPrintSemicolon
\caption{Fitness evaluation/Simulation}
\label{alg_simulation}
\KwIn{Urgency function $\Gamma(\cdot)$}
\KwOut{Average travel time of vehicles as the fitness of $\Gamma(\cdot)$}
Load traffic dataset. \;
$t \leftarrow 0$ \;
\For {each time step $t$}{
    \For{each intersection of the road network}{
        Collect the real-time data at time step $t$ on lanes of the intersection\;
        Obtain $X_i$ for each phase $s_i$ by Eq. \eqref{eq_preprocess}\;
        Identify the current most urgent phase $s^*$ based on Eq. \eqref{eq_selection} \;
        Set the signal phase of the intersection to $s^*$ \;
    }
    Switch the corresponding red and yellow lights based on changes in signal phase at each intersection, and simulate vehicles running until the next signal phase transition. \;
    $t \leftarrow t + 1$ \;
}
Calculate the average travel time and return it\; 
\end{algorithm}

\section{Experiments}
\label{sec:experiments}

\begin{table}[!t]
\footnotesize
\setlength{\tabcolsep}{24pt} 
\renewcommand{\arraystretch}{1.2} 
\caption {Parameter settings.}
\centering
\begin{tabular}{cc}
\hline
\hline
Parameter      & Value  \\ 
\hline
Population size   & 100 \\ 
The number of generations & 50 \\ 
Method of initialization & ramped-half-and-half\\ 
Initial minimum depth & 3  \\ 
Maximum depth & 8 \\
Elitism & No \\
Parent selection  & tournament selection \\
Tournament size & 3 \\
Crossover rate & 90\% \\
Mutation rate  &  10\% \\ 
\hline
\hline
\end{tabular}
\label{table_parameters}
\end{table}

In this section, we perform a series of experiments on three real-world road networks with six traffic flow datasets to evaluate our proposed method. 
The experimental results are analyzed in detail.

\subsection{Datasets}
In this paper, we use three real-world traffic data (which are frequently used in previous works \cite{wei2019colight,chen2020toward,zhang2022expression}): Dongfeng Sub-district in Jinan, Gudang Sub-district in Hangzhou, and Upper East Side in Manhattan.
The detailed descriptions are as follows:

\textbf{Jinan}: There are 12 ($3\times 4$) intersections in Dongfeng Sub-district. 
Each intersection in this road network is configured as a four-way intersection, featuring two 400-meter long road segments in the East-West direction and two 800-meter road segments in the South-North direction.
This traffic road network dataset contains three different traffic flow datasets.

\textbf{Hangzhou}: There are 16 ($4 \times 4$) intersections in Gudang Sub-district.
Each intersection is designed as a four-way intersection, featuring two 800-meter road segments in the East-West direction and two 600-meter road segments in the South-North direction. This traffic road dataset consists of two different traffic flow datasets. 

\textbf{New York}: A large-scale scenario with 196 ($28 \times 7$) intersections in Upper East Side.
Each intersection in this road network is designed as a four-way intersection, featuring two 300-meter road segments in the East-West direction and two 300-meter road segments in the South-North direction. This traffic dataset consists of one traffic flow dataset.

\begin{table*}[!t]
\footnotesize 
\renewcommand{\arraystretch}{1.2} 
\setlength{\tabcolsep}{10pt} 
\caption {The average travel time of vehicles for all compared methods in six instances.}
\centering
\begin{tabular}{c|l|c|c|c|c|c|c}
\hline\hline
\multicolumn{2}{c|}{\multirow{2}{*}{Method}} & \multicolumn{3}{c|}{Jinan}                              & \multicolumn{2}{c|}{Hangzhou}     & \multirow{2}{*}{New York} \\ \cline{3-7} 
\multicolumn{2}{c|}{}                       & \multicolumn{1}{c|}{1} & \multicolumn{1}{c|}{2} & 3 & \multicolumn{1}{c|}{1} & 2 &  \\ \hline
 & min. &514.1358 & 425.3143 & 456.3193 & 575.5565 & 563.8812 & 1582.103 \\ 
Fixed-Time \cite{webster1958traffic} & mean &514.1358 (+) & 425.3143 (+)& 456.3193 (+)& 575.5565 (+)& 563.8812 (+)& 1582.103 (+)\\ 
 & std. & 0.0 & 0.0 & 0.0 & 0.0 & 0.0 & 0.0 \\ \hline

 & min. &374.5641 & 328.9629 & 332.9403 & 365.0634 & 446.9059 & 1335.7877 \\
MP \cite{varaiya2013max} & mean &374.5641 (+)& 328.9629 (+)& 332.9403 (+)& 365.0634 (+)& 446.9059 (+)& 1335.7877 (+)\\ 
 & std. & 0.0 & 0.0 & 0.0 & 0.0 & 0.0 & 0.0 \\ \hline

        & min.  & 334.4               & 304.8     & 306.2     & 329.3    & 415.7   & 1249.4 \\
MPLight \cite{chen2020toward} & mean & 339.6599 ($\approx$)& 307.48 (+)& 313.08 (+)&331.15 (+)&427.71 (+)&1277.48 (+)\\
        & std.  &5.4376               & 2.2723    & 5.6985    &1.6187    &7.8895      &17.0731 \\ \hline
 
        & min.  & \textbf{333.1056} & \textbf{278.6506} & \textbf{287.7198} & \textbf{312.2883} & \textbf{399.8403} & \textbf{1209.7312} \\
GPLight & mean & \textbf{337.8731} & \textbf{280.2954} & \textbf{291.562}  & \textbf{314.1162} & \textbf{403.5419} & \textbf{1227.3804} \\
        & std.  & 3.4102            & 0.837             & 2.8962            & 1.4334            & 2.5603            & 14.9831 \\ \hline\hline
\end{tabular}
\begin{tablenotes}
     \item[1] Columns represent different instances and rows represent different algorithms. "$+$" denotes that the algorithm is significantly worse than GPLight, "$-$" denotes that the algorithm is significantly better than GPLight, and "$\approx$" denotes that the algorithm is comparable to GPLight. 
\end{tablenotes}
\label{table_results}
\end{table*}

\begin{table}[!t]
\footnotesize 
\renewcommand{\arraystretch}{1.2} 
\setlength{\tabcolsep}{8pt}
\caption {The gap between different compared algorithms and GPLight.}
\centering
\begin{tabular}{c|c|c|c|c}
\hline
\hline
\multicolumn{2}{c|}{\multirow{2}{*}{Instance}}    &  \multicolumn{3}{c}{Gap ($\%$)}  \\ \cline{3-5}  
 \multicolumn{2}{c|}{}& Fixed-Time & MP & MPLight \\ \hline
\multirow{3}{*}{Jinan} & 1 & 55.28 & 13.13 & 2.57 \\ \cline{2-5}
 & 2 & 34.47 & 18.03 & 10.33 \\ \cline{2-5}
 & 3 & 56.49 & 14.18 & 7.37 \\ \hline
 \multirow{2}{*}{Hangzhou} & 1 & 83.12 & 16.15 & 5.38 \\ \cline{2-5}
 & 2 & 39.75 & 10.76 & 5.99 \\ \hline
  \multicolumn{2}{c|}{New York} & 28.91 & 8.84 & 4.08 \\
\hline
\hline
\end{tabular}
\label{table_gap}
\end{table}

\subsection{Experiment Settings}
Our experiments are conducted on Cityflow$\footnote{https://cityflow-project.github.io}$, an open-source traffic simulator designed for large-scale traffic signal control \cite{zhang2019cityflow}. 
Once the traffic data is fed into the simulator, vehicles moves towards their destinations according to the environmental settings.
The simulator provides local information in an intersection to the traffic signal control method, which in turn executes the corresponding traffic signal phases.
Following convention \cite{zhang2022expression, wu2021efficient}, each green signal is succeeded by a three-second yellow signal and a two-second all-red interval to clear the intersection.
In a multi-intersection Traffic Signal Control (TSC) system, the minimum action duration serve as critical hyper-parameters and need to be consistent when establishing the baseline. We set the minimum action duration 10-second \cite{chen2020toward,zang2020metalight}.
Besides, based on the recommendations from the previous work \cite{liao2023uncertain,liao2023towards,xu2023geneticensembles, xu2023genetic}, the parameters of GPLight is represented in Table \ref{table_parameters}.

\subsection{Compared Methods}
We compare our methods with the following two categories of baseline methods, including two traditional transportation methods and a well-known deep reinforcement learning method.

\textbf{Traditional methods:}
\begin{itemize}
    \item \textbf{Fixed-Time} \cite{webster1958traffic}: each phase is allocated a fixed time interval and operates in a pre-determined cyclic schedule, without considering the actual traffic flow conditions at the intersection.
    \item \textbf{Max Pressure (MP)} \cite{varaiya2013max}: the state-of-the-art method in the transportation field, which is an adaptive control method based on pressure response. At each time step, the phase with the maximum pressure is selected, and green time is allocated to it.
\end{itemize}

\textbf{DRL method}:
\begin{itemize}
    \item \textbf{MPLight} \cite{chen2020toward}:
    a DRL-based method that integrates the advantages of previous DRL-based methods \cite{zheng2019learning,wei2019presslight} for traffic signal control.
    This method adjusts signal control by considering the current phase and traffic movement pressure as the state, while utilizing intersection pressure as the reward to optimize the coordination of traffic flows with different directions.
\end{itemize}

We refer to our proposed method as \textbf{GPLight}, which uses GP to find accurate urgency functions for traffic signal control.

\subsection{Performance Comparison} 

We tested the average travel time of vehicles for all compared algorithms using the CityFlow \cite{zhang2019cityflow} engine.
The comparison of the methods over 10 independent runs conducted on 6 instances are shown in Table \ref{table_results}, where each column represent a instance. 
The "+/$\approx$/-" indicates that the corresponding result is significantly worse than, statistically comparable to, or better than the proposed GPLight based on the Wilcoxon rank-sum test at a significance level of 0.05 with Bonferroni correction.
In Table \ref{table_results}, the best min and mean values (of the average travel time) among all the compared algorithms are highlighted. 
To better illustrate the performance gap between algorithms, we calculate the gap values using the follow formula:
\begin{equation}
    gap = \frac{f_i-f_j}{f_j}
\end{equation}
where $f_i$ represents the average value (of the average travel time) obtained by the other algorithms, $f_j$ is the average value obtained by GPLight. 
The gap values are presented in Table \ref{table_gap}.

From the results, it can be observed that Fixed-Time performs poorly. This indicates that a static signal plan is not well-suited for dynamic traffic conditions.
Compared to Fixed-Time, MP shows significant improvement, but it still falls short of learning-based methods (i.e., MPLight and GPLight).
This is attributed to the heavy reliance on over-simplified assumptions. 
In complex traffic scenarios, overly simplistic assumptions can easily lead to local optima.
It can be seen in Table \ref{table_results} that GPLight is significantly better than MPLight in all scenarios except for a relatively small scenario $\operatorname{Jinan}_1$.
Compared to MPLight, GPLight obtained better solutions in all instances (average gap 5.95$\%$).
As indicated by the minimum values in Table \ref{table_results}, the proposed GPLight consistently found the most promising solution in all instances.
This is a promising result that indicates GP can serve as a competitive counterpart to DRL in traffic signal control.
We expect this research to provide some potential inspiration for researchers in the field of traffic signal control in the future.

\begin{figure}[!t]
\centering
\includegraphics[width=\columnwidth,trim=0 0 0 0,clip]{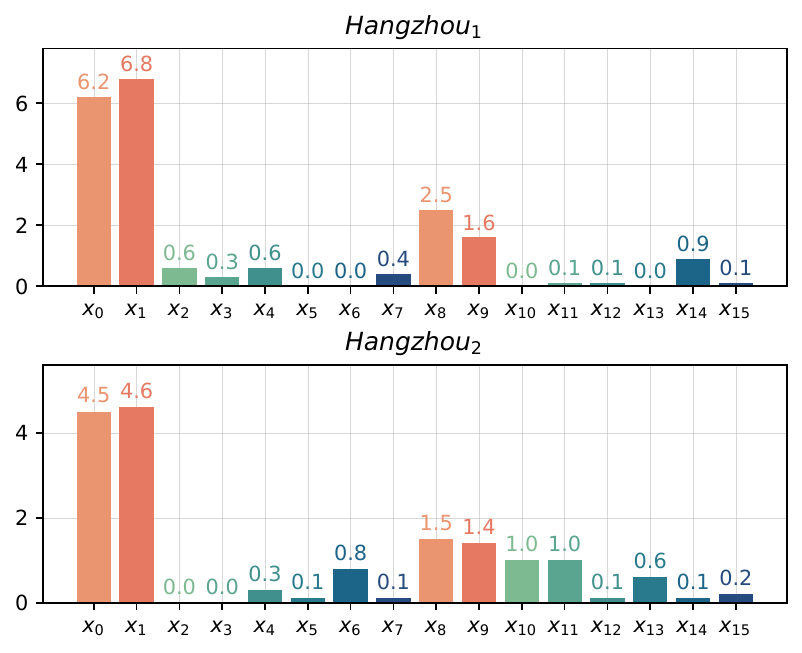}
\caption{The average frequency of occurrence of different features in the optimal solutions.}
\label{fig_terminals}
\end{figure}

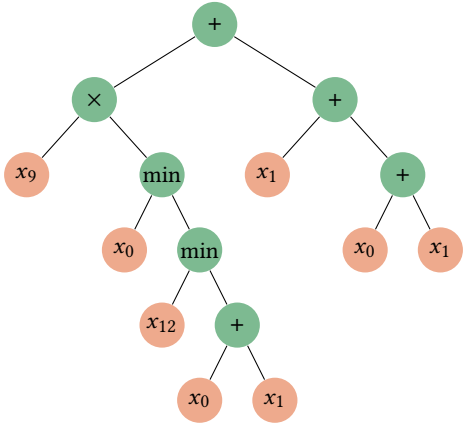
\begin{figure}[!t]
\centering
\begin{tikzpicture}
[level distance=10mm,
ternode/.style={fill=myorange!80, circle,inner sep=0pt,minimum size=6mm},
funnode/.style={fill=mygreen, circle,inner sep=0pt,minimum size=6mm},
level 1/.style={sibling distance=32mm},
level 2/.style={sibling distance=18mm},
level 3/.style={sibling distance=10mm},
my lable/.style={
rectangle split, rectangle split parts=#1, draw,rectangle split horizontal,anchor=center}
]
\node [funnode] {$+$} 
child { 
    node [funnode] {$\times$}
    child {
      node [ternode] {$x_9$} 
    }
    child {
      node [funnode] {$\min$}
      child {
        node [ternode] {$x_0$}
      }
      child {
        node [funnode] {$\min$}
        child {
          node [ternode] {$x_{12}$}
        }
        child {
          node [funnode] {$+$}
          child {
            node [ternode] {$x_{0}$}
          }
          child {
            node [ternode] {$x_{1}$}
          }
        }
      }
    }
}
child { 
    node [funnode] {$+$}
    child {
      node [ternode] {$x_1$}
    }
    child {
      node [funnode] {$+$}
      child {
        node [ternode] {$x_{0}$}
      }
      child {
        node [ternode] {$x_{1}$}
      }
    }
};
\end{tikzpicture}
\caption{An example of the evolved $\Gamma(\cdot)$ by GP. Green nodes represent functions; otherwise, they represent terminals.}
\label{fig_gpevolved}
\end{figure} 

\subsection{Feature and Rule Structure Analysis}

The evolution of symbolic expressions stands out as a notable advantage of GP. 
This often makes it easier for human to understand or explain the evolved rules.
To investigate the feature importance, we first present the average occurrence frequency of each terminal in the optimal solutions in two instances of Hangzhou in Figure \ref{fig_terminals}.
From this figure, it can be observed that $x_0$ and $x_1$, representing the number of vehicles waiting on the two incoming lanes for each phase, appear with particularly high frequencies.
If a phase is activated, the vehicles waiting on incoming lanes $l_1$ and $l_2$ for that phase are the ones most directly affected. 
Vehicles on other outgoing lanes are only indirectly influenced by that phase.
The terminals $x_8$ and $x_9$, that is, $x_{8+0}$ and $x_{8+1}$, occur with frequencies second only to $x_0$ and $x_1$, while features on outgoing lanes exhibit relatively lower occurrence frequencies.
This suggests that the features on the two incoming lanes are more important than other outgoing lanes.
And the proposed GPLight can detect important features automatically and use more important features to construct individuals.

To analyze explainability of the traffic signal control strategy, we try to analyze an example urgency function evolved by GPLight.
Figure \ref{fig_gpevolved} shows the tree structure of the urgency function trained by GPLight on $\operatorname{Hangzhou}_1$ and
its corresponding mathematical expression is:
\begin{equation}
\Gamma\left(\cdot\right) = x_{9}\min\left(x_0, \min\left(x_{12}, x_0 + x_1\right) \right) + x_1 + \left(x_0+x_1\right).
\end{equation}
Since any $x_i$ is non-negative, this urgency function can be simplified to:
\begin{equation}
  \Gamma\left(\cdot\right) = x_0 + 2x_1 + x_{8+1}\min\left(x_0, x_{8+4}\right),
\end{equation}
where $x_{8+1} = x_{9}$ and $x_{8+4} = x_{12}$.
From the first two terms of this urgency function, it can be observed that the queue lengths of the two incoming lanes for the current phase are positively correlated with the urgency of that phase. This implies that the longer the queues on the lanes involved in that phase, the higher the urgency of that phase, indicating a greater need for activation and clearance.
The last term also implies that the total number of vehicles $x_{8+1}$ on the incoming lane $l_2$ can contribute to the urgency of that phase.
This is because the total number of vehicles $x_{8+1}$ could potentially lead to an increase waiting vehicles $x_1$ in a future time.
$x_{8+1}$ also has a coefficient that is always greater than 0, which simultaneously considers the queue length of vehicles on the $l_1$ lane and the number of vehicles on the downstream go-straight lane of $l_1$.
This coefficient can adjust the contribution of $x_{8+1}$ to phase urgency based on road conditions.
GPLight is capable of obtaining an analytical expression for traffic signal control strategies based on simulation.
Compared to deep reinforcement learning, we believe that this characteristic can provide some potential insights for the explainability research in traffic signal control.

\section{Conclusions}
\label{sec:conclusion}
In this paper, we propose a new GP approach to learn better and explainable traffic signal control rules.
The experimental results indicate that using a relatively simple urgency function to evaluate the urgency for each phase achieves performance comparable to or better than a well-known DRL method.
Moreover, the proposed GPLight can provide a solution in the form of mathematical formula, which is exceedingly helpful to achieve the explainability for the traffic signal control strategy.

This work is an initial exploration, and our approach still has many shortcomings.
Although the terminal set is specially designed to address the issue of heterogeneous features, conflicts still exist among homogeneous lanes. This needs to be addressed in the future.
Additionally, this work controls traffic signals at multiple intersections in a distributed manner based on the introduced urgency function. In the future, a promising direction is to introduce communication mechanisms to further enhance the algorithm's performance.

\bibliographystyle{ACM-Reference-Format}

\appendix

\end{document}